\documentclass[letterpaper, 10 pt, conference]{ieeeconf}  

\IEEEoverridecommandlockouts                              

\usepackage{graphics} 
\usepackage{epsfig} 
\usepackage{amsmath} 
\usepackage{amssymb}  
\usepackage{mathrsfs}
\usepackage{setspace}
\usepackage{bm}
\usepackage{color}
\usepackage{epstopdf}
\usepackage[colorlinks=true,
citecolor=green,
linkcolor=red,
anchorcolor=green,
urlcolor=blue]{hyperref}     
\usepackage{url}
\usepackage{algorithm}
\usepackage{algorithmic}
\usepackage{soul} 
\soulregister{\cite}7
\soulregister{\citep}7 
\soulregister{\citet}7 
\soulregister{\ref}7 
\soulregister{\pageref}7 

\usepackage{xcolor}

\title{\LARGE \bf
Differential Flatness of Lifting-Wing Quadcopters Subject to Drag and Lift for Accurate Tracking 
}

\author{Shuai Wang, Wenhan Gao and Quan Quan
\thanks{Shuai Wang, Wenhan Gao and Quan Quan(Corresponding Author) are with the School of Automation Science and Electrical Engineering, Beihang University, Beijing 100191, China. Email: qq\_buaa@buaa.edu.cn.}
}

\begin{document}
	\begin{sloppypar}
\maketitle
\thispagestyle{empty}
\pagestyle{empty}

\begin{abstract}

In this paper, we propose an effective unified control law for accurately tracking agile trajectories for lifting-wing quadcopters with different installation angles, which have the capability of vertical takeoff and landing (VTOL) as well as high-speed cruise flight.  First, we derive a differential flatness transform for the lifting-wing dynamics with a nonlinear model under coordinated turn condition. To increase the tracking performance on agile trajectories, the proposed controller incorporates the state and input variables calculated from differential flatness as feedforward. In particular, the jerk, the 3-order derivative of the trajectory, is converted into angular velocity as a feedforward item, which significantly improves the system bandwidth.  At the same time, feedback and feedforward outputs are combined to deal with external disturbances and model mismatch. The control algorithm has been thoroughly evaluated in the outdoor flight tests, which show that it can achieve accurate trajectory tracking.
\end{abstract}

\section*{SUPPLEMENTARY MATERIAL}
Video of the paper summary and experiments is available at \url{https://youtu.be/kj3uOk2P0FQ}.
\section{INTRODUCTION}

In recent years, hybrid unmanned aerial vehicles (UAVs) have received extensive attention. 
These vehicles inherit traditional rotary-wing and fixed-wing aircraft characteristics by compensating for the gravity through rotor-induced thrust and wing-induced lift force. Integrating the advantages of both aircraft, hybrid UAVs have the capability of vertical takeoff and landing (VTOL) as well as high-speed cruise flight.

\begin{figure}
	\centering
	\includegraphics[scale=0.8]{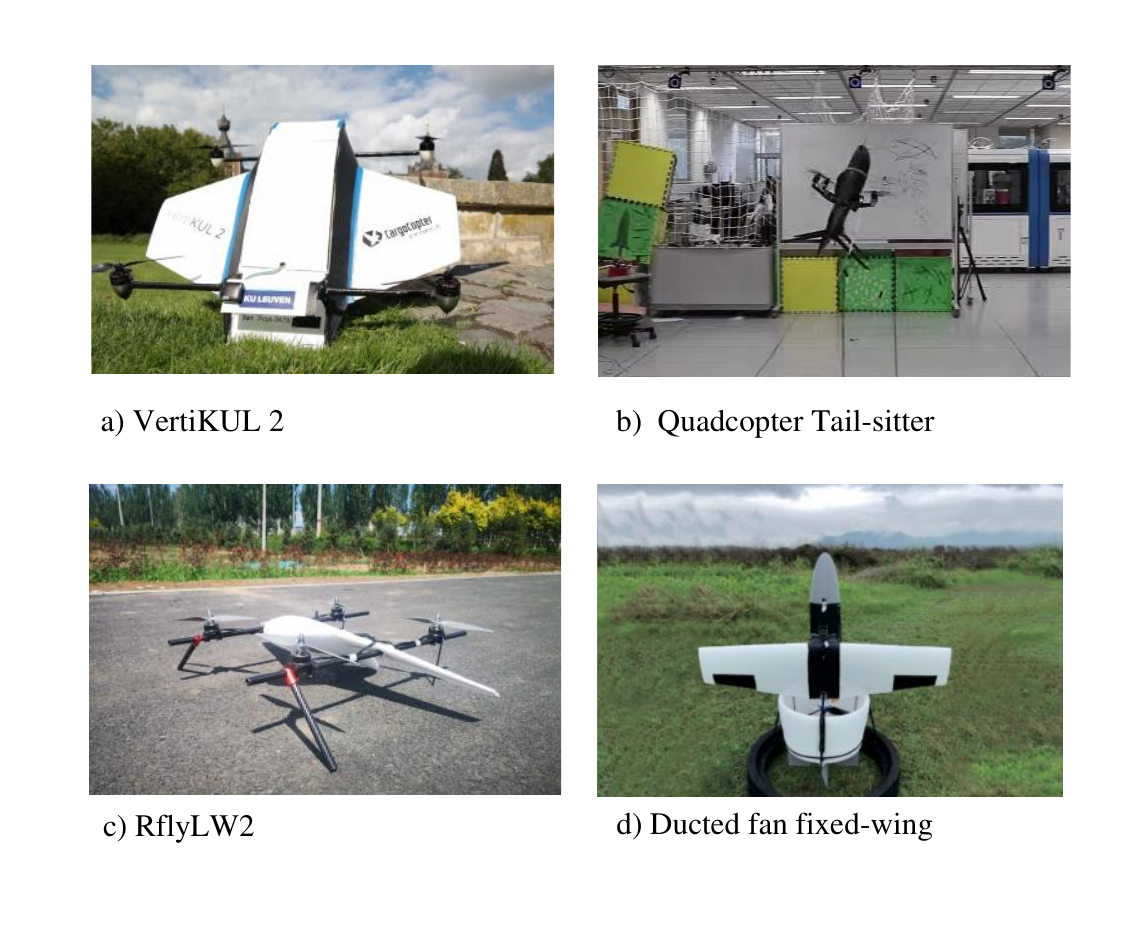}
	\caption{Prototypes of lifting-wing quadcopters. {{(a) The installation angle is 45 degrees; (b) and (d) the installation angle is 90 degrees; (c) the installation angle is designed to be 34 degrees.}}}
	\label{Fig: 1}
\end{figure}

\begin{figure}
	\centering
	\includegraphics[scale=0.9]{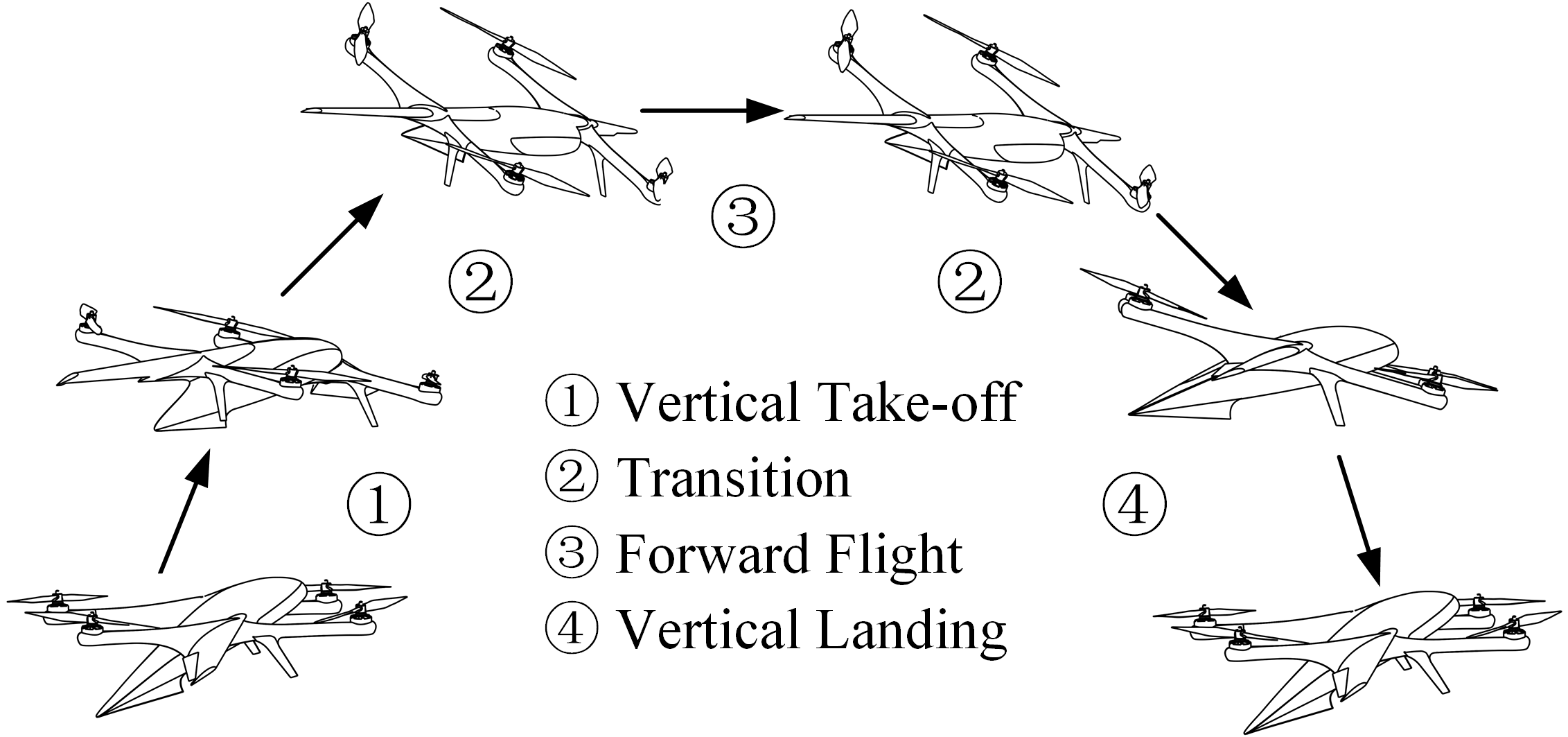}
	\caption{Different flight modes of lifting-wing quadcopters.}
	\label{Fig: transition}
\end{figure}

Among the different configurations of hybrid UAVs,  there is a type of vehicle with a structure similar to rotary-wing aircraft but with an additional wing, as shown in Fig. \ref{Fig: 1}. The characteristics of this vehicle are that the thrust direction of the rotors is at a fixed angle with the wing, and the aerodynamic interference between the rotor and the wing can be ignored
 \cite{zhang2021performance} \cite{lyu2018disturbance}. To achieve horizontal flight, the whole vehicle tilts forward using differential thrust or control vanes,  as shown in Fig. \ref{Fig: transition}. We call this type of vehicle \textit{lifting-wing quadcopters}. They have both the high maneuverability of traditional quadcopters and the long flight range of fixed-wing planes \cite{zhang2021performance}\cite{xiao2021lifting}.

Due to the underactuated property, in trajectory tracking tasks, the lifting-wing quadcopter needs to adjust its attitude to change the direction of collective thrust like quadcopters, so as to obtain the desired acceleration. Because of complex nonlinear aerodynamic forces,  the main difficulty in control design is to solve the desired attitude according to the desired acceleration (force mapping) and to make full use of the trajectory information, such as the jerk, to improve the maneuverability of the system. Aimed at these issues, {researchers have carried out extensive research}. 

Compared with traditional quadcopters, the main difference in lifting-wing quadcopters is that they have nonnegligible aerodynamic forces. {So, one method is to treat nonlinear aerodynamic forces as unmodeled disturbances. In \cite{zhou2019position}, \cite{xu2019full} and \cite{zhang2017modeling}, aerodynamic forces are modeled offline and directly added to the controller as the feedforward to minimize the model mismatch. Some studies use online estimation methods to compensate for the forces in control, such as adaptive control law \cite{cheng2022transition} and iterative learning method \cite{raj2020iterative} \cite{xu2020learning} }. In \cite{zhou2017unified}, a nonlinear optimization algorithm is used to solve the force mapping problem. However, the solver is usually too slow for real-time implementation, because the problem is nonconvex due to the complex aerodynamic model. To improve the real-time performance of the algorithm, a recurrent neural network (RNN) is trained to approximate the behavior of the nonlinear solver \cite{zhou2021control}. 

Some researchers try to solve the force mapping problem and maneuverability by utilizing the differential flatness of aircraft, whose state variables and control inputs can be written as algebraic functions of flat outputs and a finite number of their time-derivatives \cite{fliess1995flatness}. 
Using the differential flatness property, a reference trajectory generated in the flat output space can be transformed into the state-control space as feedforward terms. This has been demonstrated to improve trajectory tracking performance in agile flight \cite{faessler2017differential} and trajectory generation efficiency \cite{mueller2015computationally} for quadcopters. Only a few attempts have been made to develop the differential flatness property of the aircraft with the wing for trajectory tracking. In \cite{hauser1997aggressive}, the fixed-wing aircraft with a coordinated flight model has been shown to have the differential flatness property when thrust is considered input. However, it cannot be directly applied to aircraft with hovering capability. As for hybrid UAVs, differential flatness is derived for the tail-sitter aircraft with a simplified aerodynamics model \cite{tal2021global}, but requires an estimate of the aircraft's lateral aerodynamic force, which is often unavailable.
 
 In this paper, we derive a differential flatness transform for lifting-wing quadcopters subjected to drag and lift, when thrust and angular velocity are considered inputs, so that the force mapping problem is solved. Then, in order to accurately track the agile trajectory, we propose a unified control law, which incorporates the state and input variables calculated from differential flatness as feedforward. In particular, the jerk, the 3-order derivative of the trajectory, is converted into angular velocity as a feedforward item, which significantly improves the system bandwidth. Finally, we verified our algorithm through sufficient outdoor flight experiments, and the flight speed reached up to 10 m/s.

The remainder of this paper is organized as follows:  In Section \ref{Sec: model}, we elaborate on the aircraft dynamics with the aerodynamics model in the established coordinate system. Then, in Section \ref{Sec: differential flatness}, we derive a differential flatness transform to compute the attitude, thrust and angular velocity from the reference trajectory. In Section \ref{Sec: control law}, a controller for trajectory tracking is designed, and experimental results are shown in Section \ref{Sec: experiments}. Finally, conclusions are given in Section \ref{Sec: conclusion}.
\begin{figure}
	\centering
	\includegraphics[scale=0.90]{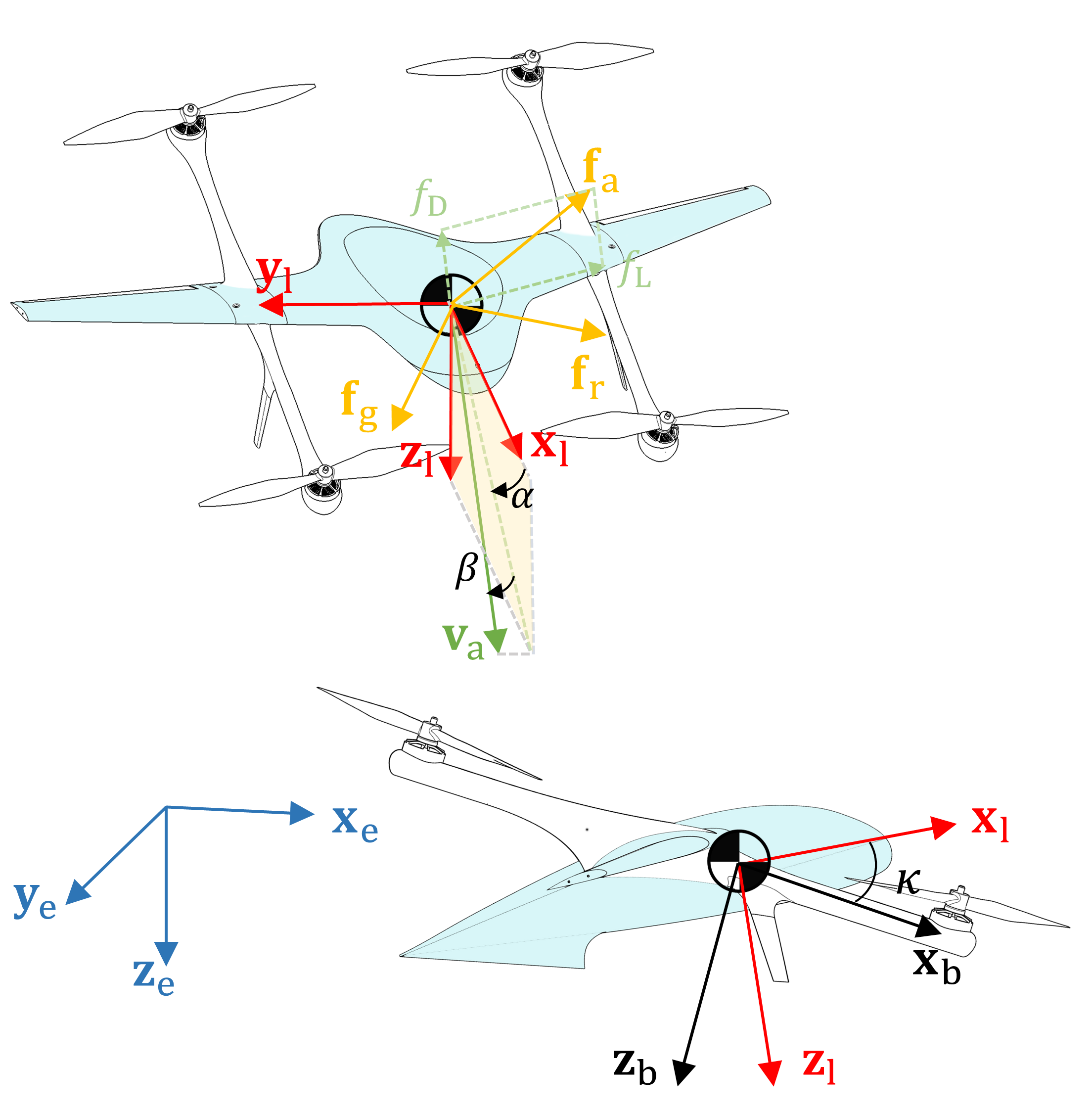}
	\caption{The lifting-wing quadcopter coordinate system and nomenclatures.}
	\label{Fig:Coordinates}
\end{figure}

\section{Problem Formulation} \label{Sec: model}
\subsection{Coordinate}
The coordinate frames of the lifting-wing quadcopter are illustrated in Fig. \ref{Fig:Coordinates}. The earth-fixed north-east-down (NED) coordinate $ {}^{\rm e}{\mathcal F} $, whose basis is composed of the column vectors of the identity matrix  $\left[ \mathbf e_{x }, \mathbf e_{y}, \mathbf e_{z} \right]$, is chosen as the inertial frame. The quadcopter-body coordinate frame ${}^{\rm b}\mathcal{F} $ is made of an orthonormal basis $\left\{\mathbf x_{\text b}, \mathbf y_{\text b} , \mathbf z_{\text  b} \right\}$ represented in ${}^{\text e}\mathcal{F} $ with the origin located at the center of mass. ${}^{\rm b}\mathcal{F} $ rotates along $\mathbf y_{\text b}$ by  $\kappa$  to obtain the lifting-wing coordinate frame ${}^{\rm l}\mathcal{F} $, which is made of orthonormal basis $\left\{\mathbf x_{\text l}, \mathbf y_{\text l} , \mathbf z_{\text  l} \right\}$ represented in ${}^{\rm e}\mathcal{F} $. The vector $\mathbf x_{\text l}$ coincides with the chord line and the lifting wing symmetry plane. To handle the wing-induced  forces, the wind coordinate frame ${}^{\rm w}\mathcal{F} $ is defined as  $\left\{\mathbf x_{\text w },\mathbf y_{\text w } ,\mathbf z_{\text  w }  \right\}$ with the vector $\mathbf x_{\text w }$ coincides with the airspeed vector $\mathbf v_{\rm a}$. The rotation matrix ${\mathbf R}_{\text{b}}^{\text{e}}=\left[ \mathbf x_{\text b}\  \mathbf y_{\text b} \ \mathbf z_{\text  b} \right] $ gives the transformation from ${}^{\rm b }\mathcal{F} $ (indicated by the subscript b) to ${}^{\rm e}\mathcal{F} $ (indicated by the superscript e). Throughout this paper, the superscripts $\left( {}^{\text {e,b,l,w}}  \mathbf{x}  \right) $ are utilized to specify in which coordinate frame a vector $\mathbf{x} $ is formulated. It should be noted that the superscripts are omitted for vectors in ${}^{\rm e}\mathcal{F} $. The position of the lifting-wing quadcopter is denoted as $\mathbf p  $, and its derivatives, velocity, acceleration and jerk as $\mathbf v$, $\mathbf a $, $\mathbf j$, respectively. 

\subsection{Lifting-wing Quadcopter Model}
The lifting-wing quadcopter is modeled as a rigid body with six degrees of freedom (6-DOF). The control inputs to the system are taken as the collective thrust $f_z$ and the angular velocity expressed in ${}^{\rm b}\mathcal{F} $ as ${}^{\rm{b }}{\bm \omega} $. It is assumed that the bandwidth of the angular velocity controller is high enough that the angular dynamics can be neglected.
The  motion of the aircraft is formulated by Newton and Euler’s Law of Motion \cite{Multicopter2017}:

\begin{align}
		\dot{{\mathbf p}} &=\mathbf{v} \label{Eq: pos kinematic} \\ 	
		\dot{{\mathbf v}} &= \mathbf{a} =\frac{\mathbf{f}}{m}  \label{Eq: pos dynamic}\\
		{\dot{\mathbf R}}_{\text{b}}^{\text{e}}&={\mathbf R}_{\text{b}}^{\text{e}}{\left[ {{}^{\text{b}}{\bm{\omega }}} \right]_ \times } \label {Eq: attitude  kinematic}
\end{align}
where $m $ is the mass of the aircraft, and ${\left[ {{}^{\text{b}}{\bm{\omega }}} \right]_ \times }$ denotes the skew-symmetric matrix. In addition, $ \mathbf f$  represents forces acting on the airframe expressed in ${}^{\rm e}\mathcal{F} $. As shown in Fig. \ref{Fig:Coordinates}, the forces acting on the aircraft are
\begin{equation}
	{\mathbf f} = {\mathbf{R}_{\rm{b }}^{\rm{e }}}\cdot{}^{\text b}{\mathbf f_{\text r}}+{\mathbf{R}_{\rm{l}}^{\rm{e}}} \cdot {}^{\rm l}{\mathbf f_{\text a}}+ m{\mathbf g}
	\label{Eq: forces}
\end{equation}
where $ {}^{\text b}{\mathbf f_{\text r}}=\left[ 0\  0\  f_{z } \right]^{\text T}$ with $f_{z }\leq 0$ is the collective thrust produced by  rotors, $ {}^{\rm l}{\mathbf f_{\text a}}$ is aerodynamic forces produced by the lifting wing, and ${\mathbf g}=\left[ 0\  0\  g \right]^{\text T}$  is the gravitational acceleration vector with $g= 9.81 {\rm m/s}^2  $.

A special parametric expression is used to model steady aerodynamic forces $ ^{\rm s}\mathbf f _{\rm a}$ \cite{pucci2011nonlinear}:
\begin{equation}
	^{\rm s }{\mathbf{f}}_{{\text{a}}} = \frac{1}{{2}}\rho S{V_a^2}\left[ {\begin{array}{*{20}{c}}
			{{C_{{d_0}}}}+C_{L \alpha}\sin ^2 \alpha  \ \ 0 \ \ C_{L \alpha}\sin \alpha \cos \alpha 
	\end{array}} \right] ^{\rm T}
\end{equation}
where $\rho $ is the air density, $S$ is the surface of the wing. The airspeed vector is defined as $\mathbf{v}_{\rm a} =\mathbf{v}-\mathbf{v}_{\rm w}$, $\mathbf{v}_{\rm w}$ is wind velocity expressed in $^{\rm e}\mathcal F$, $V_a= \lVert\mathbf {\bf v}_{\rm a}\rVert$ is the airspeed, $\alpha$ is the angle of attack; aerodynamic parameters $C_{d_0}$ is the minimum drag coefficient, $C_{y_0}$ is the minimum side force coefficient, $C_{l \alpha} $ is a coefficient related to lift, which are determined by the airfoil. This model is further developed and improved in \cite{lustosa2019global}, where a $\phi-$theory method is proposed and demonstrates that the model can cover the main dynamic characteristics in the full flight envelope.

When converted to $ {}^{\rm e}{\mathcal F} $, $ ^{\rm s}\mathbf f _{\rm a}$ is an expression independent of the angle of attack:
\begin{equation}
^{\rm l }{\mathbf{f}}_{{\text{a}}} = \frac{1}{{2}}\rho S{V_a}\left[ {\begin{array}{*{20}{c}}
		{{C_{{d_0}}}}&0&0 \\ 
		0&{{C_{{y_0}}}}&0 \\ 
		0& 0 & C_{d_0}+C_{L \alpha} 
\end{array}} \right] {{\mathbf{R}}_{\rm b}^{\rm l}} \left( {{\mathbf{R}}_{\rm b}^{\rm e}} \right)^{\rm T} \cdot {\mathbf{v}_{\rm a}}
	\label{Eq: aero forces}
\end{equation}
 Rotation matrix ${{\mathbf{R}}_{\rm b}^{\rm l}} $  transforms a vector from $ {}^{\rm b }{\mathcal F} $ to $ {}^{\rm l }{\mathcal F} $ given as 

\begin{equation}
	{{\mathbf{R}}_{\rm b}^{\rm l}}=\left[ {\begin{array}{*{20}{c}}
			\cos \kappa&0&-\sin \kappa \\ 
			0               &1&0 \\ 
			\sin \kappa&0&\cos \kappa
	\end{array}} \right]
	\label{Eq: Rbl}
\end{equation}
where $\kappa $ is the installation angle of the lifting wing. In practice applications, different installation angles can be designed according to specific flight indexes, such as flight speed and range. Generally, the installation angle is set at about 45 degrees to ensure good wind resistance \cite{zhang2021performance} \cite{theys2016control}, as shown in Fig. \ref{Fig: 1}(a) and (c). In particular, when the installation angle is set at 90 degrees, it is the so-called tail-sitter aircraft, as shown in Fig. \ref{Fig: 1}(b) and (d). This type of aircraft is more similar to the fixed wing in cruise flight  \cite{lyu2018disturbance}\cite{cheng2022transition}.  In this paper, we limit $\kappa \in (15^\circ, 90^\circ ]$, because when $\kappa$ is small, additional propellers are usually added, which is a common dual-system aircraft \cite{saeed2015review}.
\subsection{Objective}
The control design aims to accurately track the trajectory reference $ {\mathbf p_{\rm ref}}$. The reference ${\mathbf p_{\rm ref}} $, which is at least 3-order continuous,  may be provided by a pre-planned trajectory or an online motion planning algorithm.  

To facilitate the problem description, a brief introduction to differential flattening is given. Considering the following dynamic system: 
\begin{align}
	\dot{\bm X} &= \mathcal A \left( \bm X , \bm U \right)\\
	{\bm Y} &= \mathcal B \left( \bm X , \bm U \right)
	\label{Eq: dynamic system},
\end{align}
an essential property of flat systems is that their state ${\bm X}$ and input ${\bm U}$ can be directly written as algebraic functions of the flat output ${\bm Y}$ and a finite number of its time-derivatives. However, the input and state variables calculated from the flat output cannot be used to directly control the dynamic system \eqref{Eq: dynamic system} because of the inevitable model uncertainty. In practice, they can serve as feedforward control inputs to improve trajectory tracking performance by reducing the phase lag. At the same time, the model uncertainty can be compensated for by the feedback control.

In this paper, we choose ${\bm Y}= \mathbf p$ as the flat output,  ${\bm U} = \left\{ f_z,  {^{\rm b}{\bm\omega}}\right\} $ as inputs, $\bm X =\left\{ \mathbf {p,v,a,R_{\rm b}^{\rm e}} \right\}$ as states, and $\bm X $ and $\bm U $ are constrained by dynamic equations \eqref{Eq: pos kinematic}-\eqref{Eq: attitude  kinematic}. Then the objective of this paper is to address the following two problems :
\begin{itemize}
	\item Find the function of $\mathcal H$ and $\mathcal G$ that  satisfy 
	\begin{align}
			{\bm X} &= \mathcal H \left( \bm Y,\dot{\bm Y},\ldots,\bm Y^{(n)}\right) \label{Eq: differential flatness1}\\
			{\bm U} &= \mathcal G \left( \bm Y,\dot{\bm Y},\ldots,\bm Y^{(m)}\right).
		\label{Eq: differential flatness2}
	\end{align}
	\item Given a reference trajectory $\mathbf p_{\rm ref}(t)$, design a controller for the lifting-wing quadcopter with uncertainty such that $\mathop {\lim }\limits_{t \to \infty } \left\| {\mathbf p(t) - \mathbf p_{\rm ref}(t)} \right\| = 0$ by using the differential flatness property given by \eqref{Eq: differential flatness1} and \eqref{Eq: differential flatness2}. 
\end{itemize}

{The two problems will be solved in Section \ref{Sec: differential flatness} and Section \ref{Sec: control law}, respectively.}

\section{{{DIFFERENTIAL FLATNESS}}}  \label{Sec: differential flatness}

In this section, we show that the dynamics of a lifting-wing quadcopter subject to lift and drag is differentially flat. In fact, the states $\left\{ \mathbf {p,v,a,R_{\rm b}^{\rm e}} \right\}$ and inputs $\left\{ f_z, ^{\rm b}{\bm \omega} \right\} $ can be written as algebraic functions of the flat output $ {\mathbf p} $ and its derivatives. The position $\mathbf p $, velocity $\mathbf v $ and acceleration $\mathbf a$ can be easily obtained from Eqs \eqref{Eq: pos kinematic}-\eqref{Eq: attitude  kinematic}. Next, we will derive the reference attitude and thrust from the reference trajectory and its derivatives, and prove how to obtain the reference angular velocity under the coordinated turn condition.

\subsection{Attitude and Thrust } {\label {Subsec: Attitude and Thrust}}
\begin{figure}
	\centering
	\includegraphics[scale=0.95]{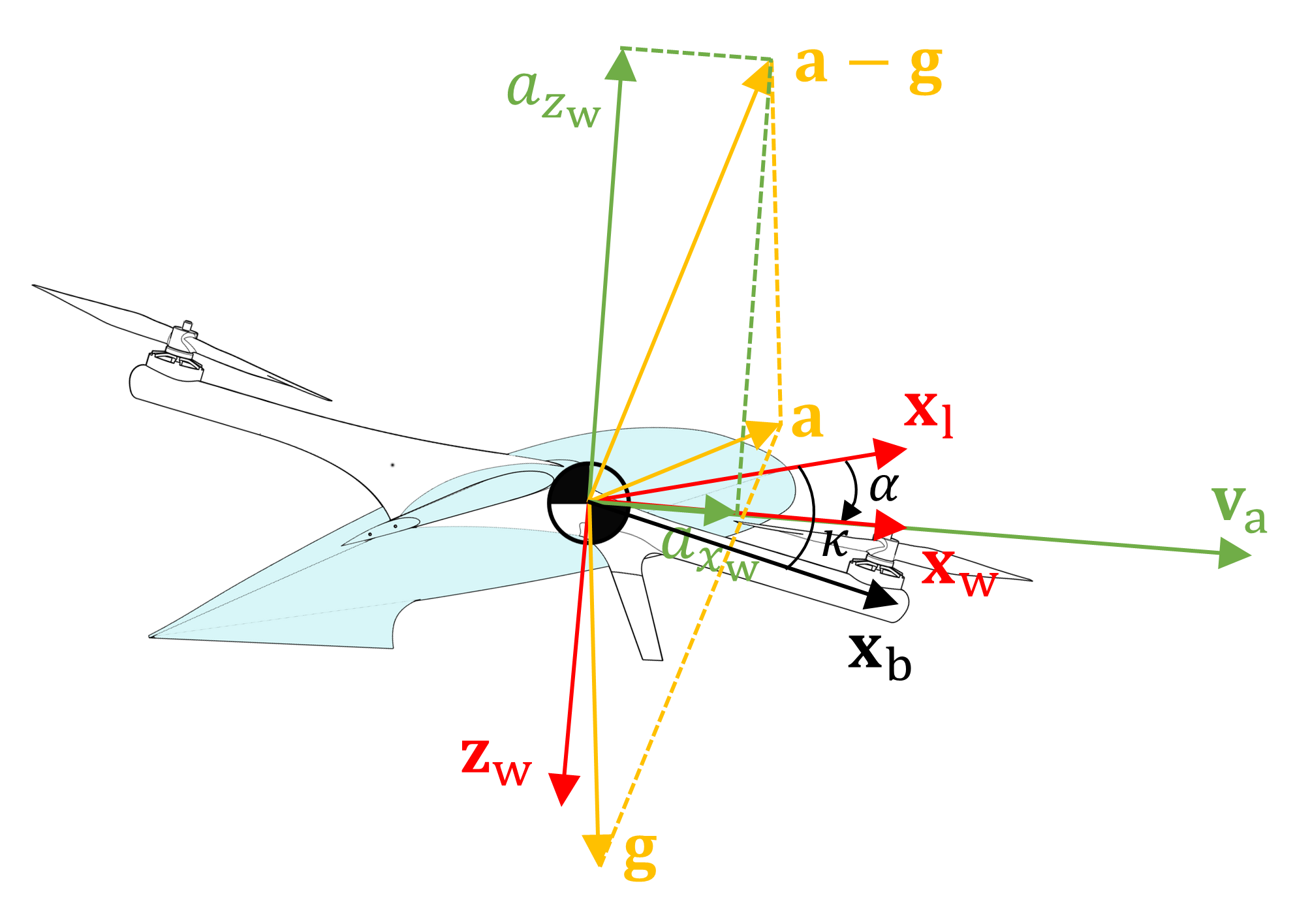}
	\caption{The reference acceleration is constrained in ${\mathbf x_{\rm w}}-{\mathbf z_{\rm w}} $ plane with the assumption that the reference sideslip angle is $0^\circ$.}
	\label{Fig: forces}
\end{figure}

 During a coordinated turn, there is no lateral acceleration expressed in $ {^{\rm w} \mathcal{F}}$ of the aircraft, that is $ ^{\rm w}{\mathbf{a}}={^{\rm w}{\mathbf f}}/ m=\left[ a_{x_{\rm w}}\ 0\ a_{z_{\rm w}}\right]^{\rm T}$, as shown in Fig. \ref{Fig: forces}. 
Then, the dynamic equation \eqref{Eq: pos dynamic} is reformulated as
\begin{equation}
	{\mathbf x}_{\rm w}{a_{x_{\rm w}}}+ {\mathbf z}_{\rm w}{a_{z_{\rm w}}} + {\mathbf g} - \mathbf{a}=0.
	\label{Eq: pos dynamic s}
\end{equation}
Since the vector $\mathbf x _{\rm w}$ coincides with the airspeed vector, $\mathbf x _{\rm w}$  is calculated as 
\begin{align}
	{\mathbf{x}}_{\rm w} = \mathbf{v} / \lVert \mathbf{v} \rVert 
	 \label{Eq: xs}
\end{align}
when the wind velocity $ \mathbf v_{\rm wind} = \bm 0$ and the norm of the reference velocity $\lVert\mathbf v_{\rm {ref}}\rVert \neq 0$. In fact, if the wind velocity can be estimated online,  the results of differential flatness hold true. For
the convenience of the following derivation, the wind velocity is assumed to be 0. The case where $\lVert\mathbf v_{\rm {ref}}\rVert =0$ will be discussed later in section \ref{Special Cases}.

Left-multiplying \eqref{Eq: pos dynamic s} by ${\mathbf x}_{\rm w}^{\rm T}$ yields
\begin{equation}
	a_{x_{\rm w}} = {\mathbf{x}}_{\rm{w}}^{\rm{T}}\left( {{\mathbf{a}} -{\mathbf g}} \right) .
	\label{Eq: axs}
\end{equation}
In $ {}^{\rm w}{\mathcal{F}}$,  the reference acceleration is constrained in ${\mathbf x_{\rm w}}-{\mathbf z_{\rm w}} $ plane. Then, the component of the reference acceleration along $ \mathbf{z}_{\rm w}$ can be derived from \eqref{Eq: pos dynamic s} by substituting \eqref{Eq: axs} and applying the Euclidean norm 
\begin{align}
	a_{z_{\rm w}} &=  - \lVert \left( {{\mathbf{I}} - {{\mathbf{x}}_{\rm w}}{\mathbf{x}}_{\rm w}^{\rm T}} \right)\left( {{\mathbf{a}} - {\mathbf{g}}} \right)\rVert \label{Eq: azs}.
\end{align}

Further, considering aerodynamic forces and rotor thrust, the transitional dynamics of the aircraft expressed in ${}^{\rm w}\mathcal{F} $ are 
\begin{equation}
	m \cdot {^{\rm w}{\bf a}}= {\mathbf R_{\rm b}^{\rm w}} \cdot {^{\rm b}{{\bf f}_{r}}} + {\mathbf R_{\rm l}^{\rm w} }\cdot{^{\rm l}{\bf f}_{a}}
	\label{Eq: as}
\end{equation}
where ${\mathbf R_{\rm b}^{\rm w} } = {\mathbf R_{\rm l}^{\rm w} } {\mathbf R_{\rm b}^{\rm l}}$ and ${\mathbf R_{\rm l}^{\rm w} }$ is the transformation from $ {}^{\rm l}{\mathcal{F}}$ to $ {}^{\rm w}{\mathcal{F}}$ given by
\begin{equation}
	{\mathbf R_{\rm l}^{\rm w} }= \left[ {\begin{array}{*{20}{c}}
			\cos \alpha&0&\sin \alpha \\ 
			0               &1&0 \\ 
			-\sin \alpha&0&\cos \alpha
	\end{array} }\right]
	\label{Eq: Rsl}.
\end{equation}

The thrust $f_z$ and angle of attack $\alpha$ are now obtained by solving the equation \eqref{Eq: as} as
\begin{align}
	f_z &=- \frac{1}{k_1} \left| k_{f_{xs}}^2 + {C_{L\alpha }}QSk_{f_{xw}} +{m^2}a_{z_{\rm w}}^2 \right| , \label{Eq: fz}\\
	\alpha  &= 2{\rm atan} \left(\frac{{\left( C_{L\alpha }QS +  k_{f_{xw}}\right)\sin \kappa  - m{a_{z_{\rm w}}}\cos \kappa - {k_1}}}{{k_{f_{xw}}\cos \kappa  + m{a_{z_{\rm w}}}\sin \kappa }}\right)
	\label{Eq: alpha}
\end{align}
where 
\begin {align}
k_{f_{xw}} &= C_{d0}QS + m{a_{x_{\rm w}}}, \notag\\
{k_1} &= \sqrt { \Big\{ \begin{aligned}{k_{f_{xs}}^2}{{\cos }^2}\kappa  &+ {{(k_{f_{xs}} + {C_{L\alpha }}QS)}^2}{{\sin }^2}\kappa  + {m^2}a_{z_{\rm w}}^2\\
		&-m{C_{L\alpha }}QS{a_{z_{\rm w}}}\sin 2\kappa \end{aligned}\Big\}}.\notag
\end{align}

Substituting  \eqref{Eq: forces} and \eqref{Eq: aero forces} into \eqref{Eq: pos dynamic}, the position dynamic equation is reformulated as
\begin{align}
	{c_z}{{\mathbf{z}}_{\rm b}}& - \left( {{Q_a}{C_{{d_x}}}{\mathbf{x}}_{\rm b}^{\rm T}{{\mathbf{v}}} + {Q_a}{C_{{d_{xz}}}}{\mathbf{z}}_{\rm b}^{\rm T}{\mathbf{v}}} \right){{\mathbf{x}}_{\rm b}} - 
	\left( {{Q_a}{C_{{y_0}}}{\mathbf{y}}_{\rm b}^{\rm T}{\mathbf{v}}} \right){{\mathbf{y}}_{\rm b}}  \notag \\
	&- \left( {{Q_a}{C_{{d_z}}}{\mathbf{z}}_{\rm b}^{\rm T}{\mathbf{v}} + {Q_a}{C_{{d_{xz}}}}{\mathbf{x}}_{\rm b}^{\rm T}{\mathbf{v}}} \right){{\mathbf{z}}_{\rm b}} + g{{\mathbf{z}}_{\rm e}} - {\mathbf{a}} = 0
	\label{Eq: pos dynimic1}
\end{align}
where  ${C_{{d_x}}} = {C_{{d_0}}}{\cos ^2}\kappa  + ({C_{{L_\alpha }}} + {C_{{d_0}}}){\sin ^2}\kappa $, ${C_{{d_z}}} = {C_{{d_0}}}{\sin ^2}\kappa  + ({C_{{L_\alpha }}} + {C_{{d_0}}}){\cos ^2}\kappa $, ${C_{{d_{xz}}}} = {C_{{L_\alpha }}}\sin \kappa \cos \kappa $, ${c_z} = {{{f_z}} \mathord{\left/{\vphantom {{{f_z}} m}} \right.\kern-\nulldelimiterspace} m}$,  ${Q_a} = {{\rho S{V_a}} \mathord{\left/{\vphantom {{\rho S{V_a}} {2m}}} \right.\kern-\nulldelimiterspace} {2m}}$.  

Left-multiplying  \eqref{Eq: pos dynimic1} by  ${\mathbf{y}}_{\rm b}^{\rm T}$ yields
\begin{equation}
	{\mathbf{y}}_{\rm b}^{\rm T}{{\mathbf{y}}_{\rm {perp}}} = 0
	\label{Eq: yb constrain}
\end{equation}
with
\begin{equation}
	{{\mathbf{y}}_{ \rm perp}} = {Q_a}{C_{{y_0}}}{\mathbf{v}} - g{{\mathbf{z}}_e} + {\mathbf{a}}.
	\label{Eq: yperp}
\end{equation}

Under the coordinated turn condition,  $\mathbf x_{\rm w} $ can be obtained by rotating  $\mathbf x_{\rm b} $ along  $\mathbf y_{\rm b} $ by $\alpha -\kappa$, which implies that $\mathbf y_{\rm b} $ is perpendicular to $\mathbf x_{\rm w} $. With this and \eqref{Eq: yb constrain}, $\mathbf y_{\rm b} $ is obtained as
\begin{equation}
	{{\mathbf{y}}_{\rm b}} = \frac{{{\mathbf{x}}_{\rm w}} \times {{\mathbf{y}}_{ \rm perp}} }{{\left\| {{{\mathbf{x}}_{\rm w}} \times {{\mathbf{y}}_{ \rm perp}}} \right\|}}.
	\label{Eq: yb}
\end{equation}
According to Rodrigues' formula, $\mathbf x_{\rm b} $ is obtained as
\begin{align}
	{{\mathbf{x}}_{\rm b}} = \big(& \cos \left( {\alpha  - \kappa } \right){\mathbf I} + \left( {1 - \cos\left( {\alpha  - \kappa } \right)} \right){\mathbf{y}}_{\rm b}{\mathbf{y}}_{\rm b}^{\rm T} \notag \\ 
	&	+ \sin \left( {\alpha  - \kappa } \right) {\left[ {\mathbf y}_{\rm b} \right]_ \times } \big){{\mathbf{x}}_{\rm w}}.
	\label{Eq: xb}
\end{align}

Then, utilizing the orthogonality of rotation matrix, $\mathbf R_{\rm b}^{\rm e} $  is constructed as
\begin{equation}
	\mathbf R_{\rm b}^{\rm e} = \left[ {\mathbf x}_{\rm b}\ {{\mathbf{y}}_{\rm b}}\ {{\mathbf{x}}_{\rm b}} \times {{\mathbf{y}}_{\rm b}}\right]
	\label{Eq: R_eb}.
\end{equation}

\subsection{Angular Velocity} {\label {Subsec: Angular Velocity}}
Taking the derivative of \eqref{Eq: pos dynimic1} yields
	\begin{align}
	{\mathbf{j}} =& {\dot c_z}{{\mathbf{z}}_{\rm b}} + {c_z}{{\mathbf{\dot z}}_{\rm b}} - {Q_a}\mathbf R_{\rm b}^{\rm e}\left( {{\left[ {{}^{\text{b}}{\bm{\omega }}} \right]_ \times }{\mathbf D} - {\mathbf D}{\left[ {{}^{\text{b}}{\bm{\omega }}} \right]_ \times }} \right){\left( {\mathbf R_{\rm b}^{\rm e}} \right)^{\rm T}}{\mathbf{v}} \notag \\ 
	 &- 2{Q_a}\mathbf R_{\rm b}^{\rm e}{\mathbf D}{\left( {\mathbf R_{\rm b}^{\rm e}} \right)^{\rm T}}{\mathbf{a}}
	 \label{Eq: pos dynimic2}
	\end{align}
where 
\begin{equation*}
	{\mathbf D} = \left[ {\begin{array}{*{20}{c}}
			{{C_{{d_x}}}}&0&{{C_{{d_{xz}}}}} \\ 
			0&{{C_{{y_0}}}}&0 \\ 
			{{C_{{d_{xz}}}}}&0&{{C_{{d_z}}}} 
	\end{array}} \right].
\end{equation*}
Left-multiplying  \eqref{Eq: pos dynimic2} by  ${\mathbf{x}}_{\rm b}^{\rm T}$ yields 
\begin{equation}
	\begin{gathered}
		{\mathbf{x}}_{\rm b}^{\rm T}{\mathbf{j}} + {Q_a} {{\mathbf v^{\rm T} \mathbf a}/ V_a^2} \left( {{d_x}{\mathbf{x}}_{\rm b}^{\rm T}{\mathbf{a}} + {d_{xz}}{\mathbf{z}}_{\rm b}^{\rm T}{\mathbf{a}}} \right)   \hfill \\ 
		 = {\omega _{y_{\rm b}}}\left( {{c_z} - {Q_a}\left( {\left( {{d_z} - {d_x}} \right){\mathbf{z}}_{\rm b}^{\rm T}{\mathbf{v}} + 2{d_{xz}}{\mathbf{x}}_{\rm b}^{\rm T}{\mathbf{v}}} \right)} \right)  \\
		+{\omega _{x_{\rm b}}}{Q_a}{d_{xz}}{\mathbf{y}}_{\rm b}^{\rm T}{\mathbf{v}} -{\omega _{z_{\rm b}}}{Q_a}\left( {{d_x} - {C_{{y_0}}}} \right){\mathbf{y}}_{\rm b}^{\rm T}{\mathbf{v}}.
	\end{gathered}
	\label{Eq: x_b *j}
\end{equation}
Left-multiplying  \eqref{Eq: pos dynimic2} by  ${\mathbf{y}}_{\rm b}^{\rm T}$ yields 
\begin{align}
&{\mathbf{y}}_{\rm b}^{\rm T}{\mathbf{j}} + {Q_a}\left( {{\mathbf v^{\rm T} \mathbf a}/ V_a^2}\right) {C_{{y_0}}}{\mathbf{y}}_{\rm b}^{\rm T}{\mathbf{a}}   \notag \\
&= {\omega _{x_{\rm b}}}\left( { - {c_z} + {Q_a}\left( {{d_{xz}}{\mathbf{x}}_{\rm b}^{\rm T}{\mathbf{v}} + \left( {{d_z} - {C_{{y_0}}}} \right){\mathbf{z}}_{\rm b}^{\rm T}{\mathbf{v}}} \right)} \right)  \notag \\
&- {\omega _{z_{\rm b}}}{Q_a}\left( {\left( {{d_x} - {C_{{y_0}}}} \right){\mathbf{x}}_{\rm b}^{\rm T}{\mathbf{v}} + {d_{xz}}{\mathbf{z}}_{\rm b}^{\rm T}{\mathbf{v}}} \right) .   
	\label{Eq: y_b *j}
\end{align}

Under the coordinated turn condition, there is also no lateral acceleration expressed in $ {^{\rm b} \mathcal{F}}$ of the aircraft, that is
\begin{equation}
	{\mathbf y}_{\rm e}^{\rm T}\cdot {}^{\rm b}{\mathbf{\dot v}} = 0
	\label{Eq: wbz}.
\end{equation}
With \eqref{Eq: wbz} and \eqref{Eq: pos dynamic},  another constraint on angular velocity is obtained 
\begin{equation}
	{\omega _{x_{\rm b}}}{v_{{z_{\rm b}}}} - {\omega _{z_{\rm b}}}{v_{{x_{\rm b}}}} =  - { g_{{y_{\rm b}}}}
	\label{Eq: wb constraint}.
\end{equation}

The angular velocity can now be obtained by solving the linear equations composed of \eqref{Eq: x_b *j}, \eqref{Eq: y_b *j}  and \eqref{Eq: wb constraint}.

\setlength{\baselineskip}{12pt}

\subsection{ Special Cases } \label {Special Cases}
In this section, we will introduce some practical solutions to solve the special case that the proposed method is invalid.

\textbf{Case 1 - $\lVert \mathbf v \rVert = 0 $:}
For the lifting-wing quadcopter, the case often occurs. For example, when the aircraft is commanded to hover, the reference velocity $\mathbf v_{\rm {ref}}$ is desired to be $\mathbf 0$. In fact, when ${\mathbf v }\to {\mathbf 0}$, ${{\mathbf{x}}_{\rm w}}$ and $\alpha $ are not defined, but this does not mean that ${{\mathbf{y}}_{\rm b}}$ and ${{\mathbf{x}}_{\rm b}}$ cannot be solved.
In Section \ref{Subsec: Attitude and Thrust}, we use the wind speed vector to determine the nose orientation of the aircraft, but when the wind speed vector is 0, the nose orientation of the aircraft cannot be determined accordingly. In this case, $\mathbf y_{\rm b}$, originally determined by \eqref{Eq: yb}, can be any vector on the plane perpendicular to $\mathbf y_{\rm perp}$. To determine the attitude $\mathbf R_{\rm b}^{\rm e}$ of the aircraft, the yaw angle $\psi$ should be given when $\mathbf v_{\rm {ref}} = 0$, then $\mathbf y_{\rm b}$ is obtained as 
 \begin{equation}
 	{{\mathbf{y}}_{\rm b}} = \frac{{{\mathbf{x}}_{\rm c}} \times {{\mathbf{y}}_{ \rm perp}} }{{\left\| {{{\mathbf{x}}_{\rm c}} \times {{\mathbf{y}}_{ \rm perp}}} \right\|}}
 	\label{Eq: yb when v=0}
 \end{equation}
where ${{\mathbf{x}}_{\rm{c}}} = {[\begin{array}{*{20}{c}}
		{\cos \psi }&{\sin \psi }&0 
	\end{array}]^{\rm {T}}}$. 

Left-multiplying  \eqref{Eq: pos dynimic1} by  ${\mathbf{x}}_{\rm b}^{\rm T}$ yields
\begin{equation}
	{\mathbf{x}}_{\rm b}^{\rm T}\left( {Q_a}{C_{{d_x}}}{\mathbf{v}} - g{{\mathbf{z}}_{\rm e}} + {\mathbf{a}} \right) = {Q_a}{C_{{d_{xz}}}}{\mathbf{z}}_{\rm b}^{\rm T}{\mathbf{v}}.
	\label{Eq: xb constrain}
\end{equation}
That  is, when ${\mathbf v }= {\mathbf 0}$, ${\mathbf{x}}_{\rm b}^{\rm T} {{\mathbf{y}}_{ \rm perp}}={\mathbf 0}$. With this and \eqref{Eq: yb constrain}, $\mathbf x_{\rm b} $ is obtained as
\begin{equation}
	{{\mathbf{x}}_{\rm b}} = \frac{{{\mathbf{y}}_{\rm perp}} \times {{\mathbf{y}}_{ \rm b}} }{{\left\| {{{\mathbf{y}}_{\rm perp}} \times {{\mathbf{y}}_{ \rm b}}} \right\|}}.
	\label{Eq: xb1}
\end{equation}

However, state mutation may occur by directly giving an arbitrary yaw angle. Another simple and effective method is to keep $\mathbf x_{\rm w} $ to the last state when  $ \mathbf v$ is not equal to 0.

\textbf{Case 2 - ${{\mathbf{x}}_{\rm w}} \times {{\mathbf{y}}_{ \rm perp}}  = 0 $:}
This case occurs if ${{\mathbf{y}}_{ \rm perp}}  $ is aligned with ${{\mathbf{x}}_{\rm w}}$, for example when the aircraft needs to execute the vertical takeoff or landing command. In this case, $\mathbf y_{\rm b}$ can be any vector on the plane perpendicular to $\mathbf y_{\rm perp}$. To overcome this singularity, $\mathbf y_{\rm b}$ can be obtained using the same method as in \textbf{Case 1}.

\section{CONTROL LAW}  \label{Sec: control law}
To accurately track a reference trajectory, a controller, consisting of feedback terms computed from tracking errors as well as feedforward terms $\left\{ \mathbf {p_{\rm ref},v_{\rm ref},a_{\rm ref},R_{\rm ref}},\bm {\omega}_{\rm ref}\right\}$ computed from the reference trajectory using the differential flatness property of the lifting-wing quadcopter, is designed. The control architecture consists of an outer-loop position controller and an inner-loop attitude controller. The position 
controller computes the desired acceleration, which is mapped to the collective thrust and attitude. The attitude command is sent to the inner loop, and the attitude controller generates the desired angular velocity. A high-bandwidth controller is used to track these angular velocity commands and thrust.
\begin{figure}
	\centering
	\includegraphics[scale=0.95]{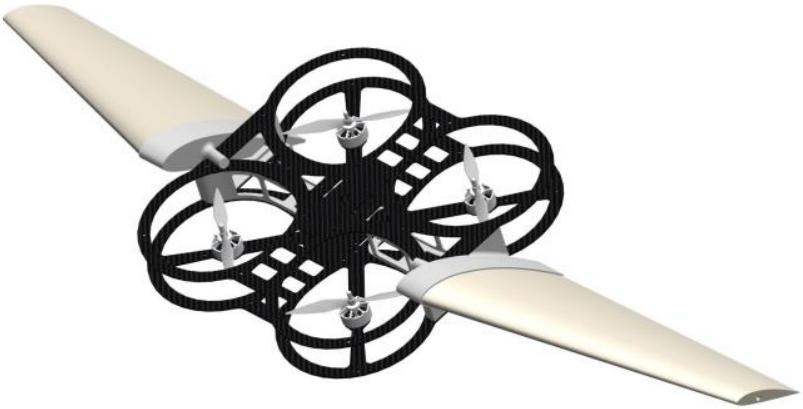}
	\caption{Lifting-wing quadcopters with a special installation angle $\kappa = 34 \deg $ for experiments.}
	\label{Fig: Lifting-wing quadcopters}
\end{figure}
\begin{figure*}[ht]
	\centering
	\includegraphics[scale=0.96]{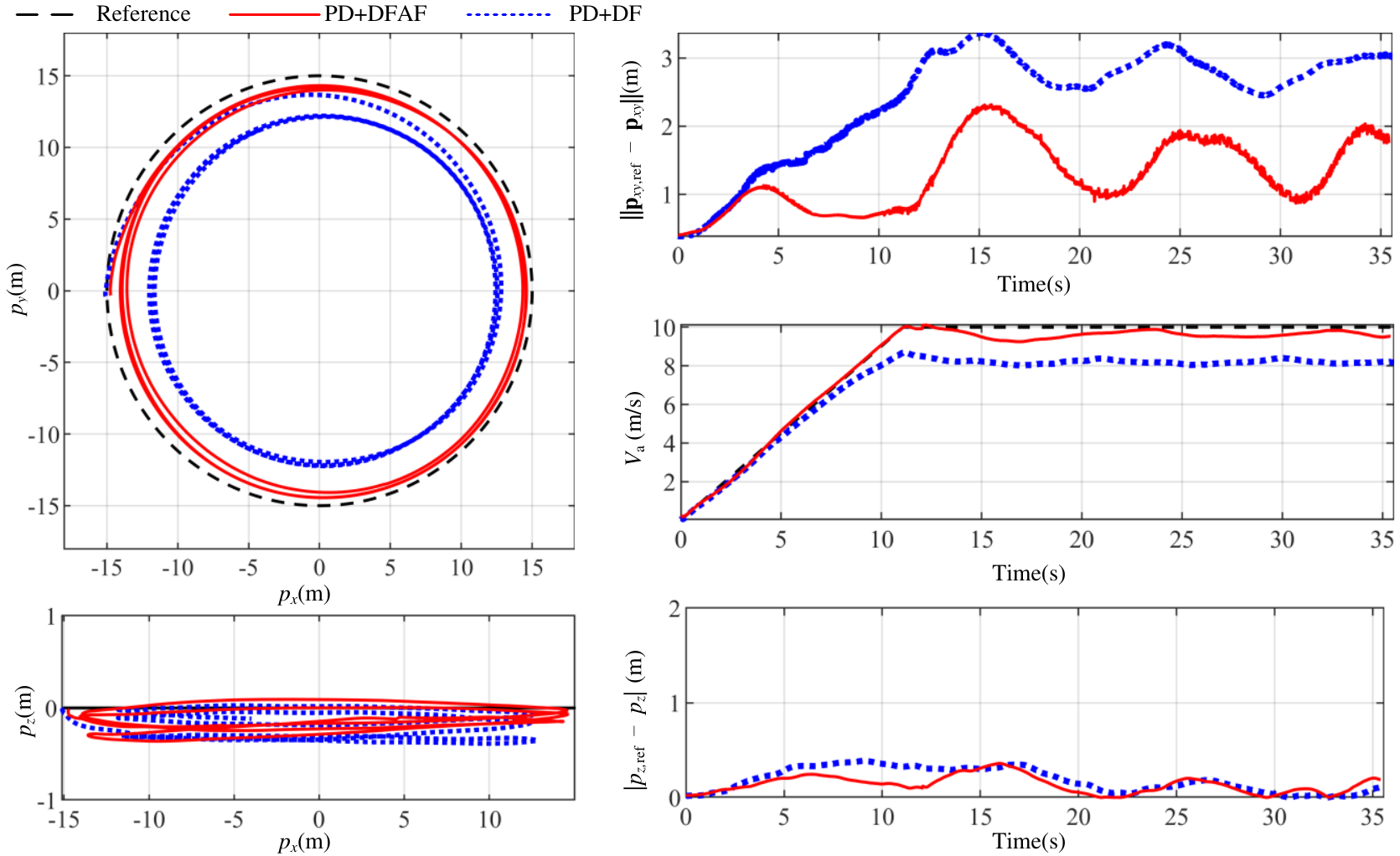}
	\caption{Experimental results for circle trajectory. Estimated local position on the circle trajectory considering aerodynamic forces (solid red), and without considering aerodynamic forces (solid blue) compared to the desired position (dashed black).}
	\label{Fig: Circular Trajectory}
\end{figure*}
\subsection{Position and Velocity Controller}
Cascaded PID controllers for position \eqref{Eq: pos kinematic} and velocity \eqref{Eq: pos dynamic} are designed as
\begin{equation}
\mathbf v_{\rm d} = {\mathbf K_{\mathbf p p}}\left( \mathbf p_{\rm {ref}} - \mathbf p  \right) + \mathbf v_{\rm{ref}},
	\label{Eq: pos controller}
\end{equation}
\begin{equation}
	\mathbf a_{\rm d} = {\mathbf K_{\mathbf v p}}\left( \mathbf v_{\rm {d}} - \mathbf v \right) +{\mathbf K_{\mathbf v i}}\int \left( \mathbf v_{\rm {d}} - \mathbf v \right) {\rm d}t +\mathbf a_{\rm{ref}}
	\label{Eq: vel controller}
\end{equation}
where ${{\mathbf{K}}_{{\mathbf{p }}{\text{p}}}}$, ${{\mathbf{K}}_{{\mathbf{v }}{\text{p}}}}$, ${{\mathbf{K}}_{{\mathbf{v }}{i}}}  \in {\mathbb{R}^{3 \times 3}}$ are diagonal matrix acting as the feedback control gain. The integral term is added to compensate for model errors and external disturbances.  According to the results of Section \ref{Sec: differential flatness}, the desired collective thrust $f_{z,{\rm d}}$ and desired attitude $\mathbf R_{\rm d}$ can be obtained from $\bf a_{\rm d}$. However, in the actual flight, there is the inevitable deviation between the aerodynamic force calculated by \eqref{Eq: aero forces} and acting on the aircraft. Multiplying the aerodynamic force by an attenuated feedforward coefficient can reduce the adverse effects caused by the inaccurate model. So the  desired mass-normalized collective thrust expressed in $^{\rm e}{\mathcal F}$ is obtained as 
\begin{equation}
	\mathbf a_{\rm r,d} = \mathbf a_{\rm{d}} -{\mathbf g}- {\mathbf K_{\rm ff}}\mathbf a_{\rm{af}}
	\label{Eq: azd}
\end{equation}
where ${\mathbf K_{\rm ff}} \in {\mathbb{R}^{3 \times 3}}$ are diagonal matrix acting as the feedforward control gain, and $\mathbf a_{\rm{af}}$ is the acceleration due to aerodynamic forces, obtained as 
\begin{equation}
	{{\mathbf{a}}_{{\rm{af}}}} = \frac{1}{{2m}}\rho S{V_a}{{\mathbf{R}}_{{\rm{ref }}}} \mathbf D {\mathbf{R}}_{{\rm{ref}}}^{\rm T}{\mathbf{v}_{\rm ref}}.
	\label{Eq: ard }
\end{equation}

The  desired mass-normalized collective thrust $\mathbf a_{\rm{r,d}} $ is completely produced by rotor thrust, so the desired attitude $\mathbf R_{\rm d}$ is obtained as
\begin{equation}
	\begin{gathered}
		{{\mathbf{z}}_{\rm{b,d}}} =  - \frac{{{{\mathbf{a}}_{\rm{r,d}}}}}{{\left\| {{{\mathbf{a}}_{\rm{d}}}} \right\|}} \hfill \\
		{{\mathbf{y}}_{\rm{b,d}}} = \frac{{{{\mathbf{z}}_{\rm{b,d}}} \times {{\mathbf{x}}_{\rm{w}}}}}{{\left\| {{{\mathbf{z}}_{\rm{b,d}}} \times {{\mathbf{x}}_{\rm{w}}}} \right\|}} \hfill \\
		{{\mathbf{x}}_{\rm{b,d}}} = {{\mathbf{y}}_{\rm{b,d}}} \times {{\mathbf{z}}_{\rm{b,d}}} \hfill \\ 
	\end{gathered}
	\label{Eq: Rb control }
\end{equation}
where $\mathbf x_{\rm w}$ is obtained by \eqref{Eq: xs}. By projecting the desired acceleration onto the quadcopter-body $z $-axis, the collective thrust input is computed as 
\begin{equation}
f_{z,{\rm d}}=m\left( \mathbf z_{\rm b,d}^{\rm T}\cdot \mathbf a_{\rm d}\right).
\end{equation}

\subsection{Attitude Controller}
The attitude error is presented in the form of quaternion based on which the corresponding controller is designed:
\begin{equation}
	{\mathbf{q}_{\rm err}} = {\mathbf{q}_{\rm d}^{*}} \otimes {\mathbf{q_{\rm b}^{\rm e}}} = {\left[ q_{\rm{e}_0}\ \ {q_{\rm{e}_1}}\ \ 
		{q_{\rm{e}_2}}\ \ {q_{\rm{e}_3}} \right]^{\rm{T}}}
\end{equation}
where ${\mathbf{q}}_{\rm{d}}$ is transformed from $\mathbf R_{\rm d}$ , $ (\cdot)^*$ is the conjugate of a quaternion, and $\otimes $ is the quaternion product. Then, ${\mathbf{q}}_{\rm{err}}$ is transformed into the axis-angle form ${\mathbf{q}}_{\rm{err}}= \left[ \cos \frac{\vartheta}{2} \ \ {\bm \xi}^{\rm T}_{\rm e}\sin \frac{\vartheta}{2} \right]^{\rm T}$ by 
\begin{equation}
	\begin{gathered}
		\vartheta  = {\rm {wrap}}_\pi \left( {2{{\arccos}}\left( q_{\rm{e}_0} \right)} \right) \hfill \\
		{\bm \xi}_{\rm{e}}  = 
				{\rm{sign}}(q_{\rm{e}_0}) \frac{\vartheta }{{\sin {\vartheta  \mathord{\left/{\vphantom {\vartheta  2}} \right.
								\kern-\nulldelimiterspace} 2}}}{{\left[ {q_{\rm{e}_1}\ \ q_{\rm{e}_2}\ \ q_{\rm{e}_3} }\right]}^{\rm{T}}}.
	\end{gathered}
\end{equation}
The function ${\rm {wrap}_\pi }(\vartheta)$  constrains the $ \vartheta$ in $\left[{-\pi\ \ \pi}\right]$ to ensure the shortest rotation path. To eliminate the attitude error, the attitude controller is designed as 
\begin{equation}
	{}^{\rm{b}}{{\bm{\omega }}_{\rm{d}}} =  {{\mathbf{K }}_{{\mathbf{\Theta }}{\rm{p}}}}{{\bm \xi} _{\rm e}} + {{ \bm\omega _{\rm {ref}}}}
\end{equation}
where ${{\mathbf{K}}_{{\mathbf{\Theta }}{\rm{p}}}}  \in {\mathbb{R}^{3 \times 3}}$ is diagonal matrix acting as the control gain.

\section{REAL-WORLD EXPERIMENTS}  \label{Sec: experiments}
In this section, we evaluate the performance of the proposed controller in outdoor scenes, including high-speed and agile circle and lemniscate trajectories.
\begin{figure}[t]
	\centering
	\includegraphics[scale=1.0]{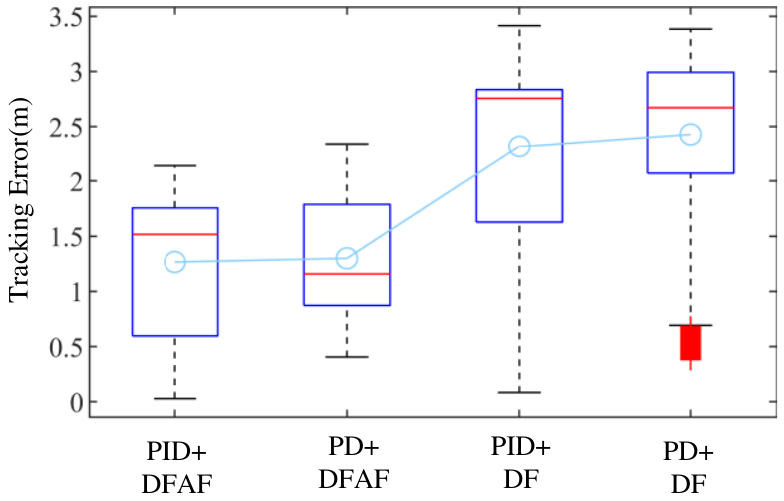}
	\caption{ The box plots of the root square position error for circle trajectory tracking. The sky blue line marked with 'o' indicates the root mean square position error.}
	\label{Fig: Compare}
\end{figure}

\begin{figure}[h]
	\centering
	\includegraphics[scale=1.0]{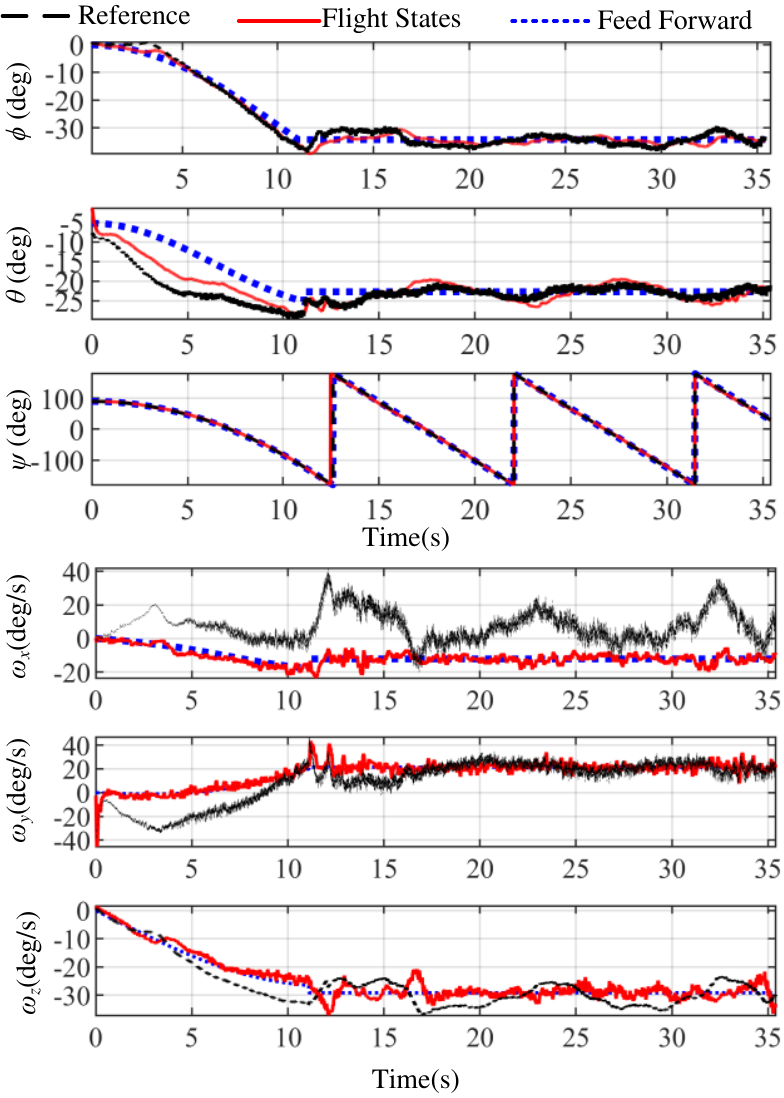}
	\caption{ Estimated attitude on the circle trajectory (solid red), the desired attitude in attitude controller (dashed black), and feedforward references computed from differential flatness transform considering aerodynamic forces (solid blue).}
	\label{Fig: circle_feedforward}
\end{figure}
\subsection{Experimental Setup}

We prepared a lifting-wing quadcopter with a special installation angle for experiments, as shown in Fig. \ref{Fig: Lifting-wing quadcopters}. The lifting wing is made of foam and connected to the quadcopter through a 3D-printing connector which controls the installation angle. The power system involves a 3300mAh, four-cell lithium polymer (LiPo) battery, EMAX RS2205 2600KV brushless DC motors with HQ5045 BN propeller, and 45A electronic speed control (ESC). A Pixhawk flight controller\footnote{https://docs.px4.io/master/en/assembly/quick\_start\_pixhawk4.html} is rigidly attached to the center of the quadcopter frame, which has an on-board inertial measurement unit (IMU), barometer and SD card for flight data logging. A separate NEO-M8N based GPS module with a magnetometer is used to read the positioning and heading. A Benewake TFmini Lidar with a 40 m range is used for altitude estimation. A 900MHz telemetry link performs communication between the ground control station and the Pixhawk. The flight control algorithm runs on the Pixhawk at 250 Hz. 

To validate our theoretical results, we present two trajectories, that is, circle and Lemniscate trajectories, and use the root mean square position error (RMSE) $E_p$ to evaluate the trajectory tracking performance, which is defined as
\begin{equation}
	{E_p} = \sqrt {\frac{1}{N}\sum\limits_{k = 1}^N {\left( {{{\bf{p}}_{{\rm{ref}}}}\left( k \right) - {\bf{p}}\left( k \right)} \right)} }
\end{equation}
where $N$ is the control cycles of the position controller required to execute a given trajectory. 
\begin{figure*}[t]
	\centering
	\includegraphics[height=10.5cm]{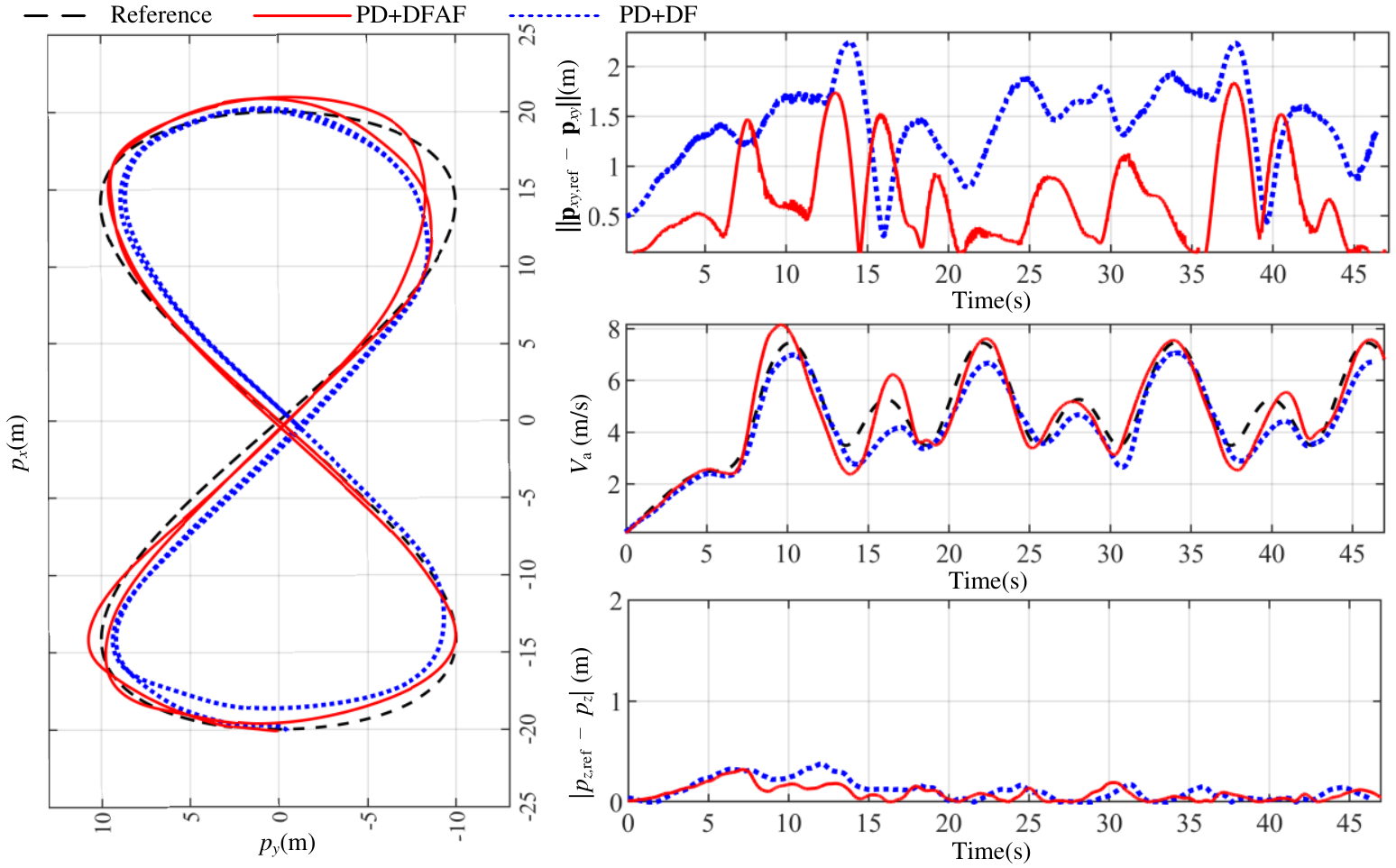}
	\caption{Experimental results for Lemniscate trajectory. Estimated local position on the Lemniscate trajectory considering aerodynamic forces (solid red), and without considering aerodynamic forces (solid blue) compared to the desired position (dashed black).}
	\label{Fig: Lemniscate Trajectory}
\end{figure*}
\subsection{Circle Trajectory}
The circle trajectory is expressed as
\begin{equation}
	\mathbf p_{\rm ref} = \mathbf p_{0}+\left[ r(1-\cos(0.5\omega t^2))\ r\sin(0.5\omega t^2)\ 0\right]^{\rm T}.
\end{equation}
where $\mathbf p_0$ is the local position of the vehicle when the task starts. Here, we set $r=15$ m, $\omega = 0.06$ rad/s, and when the magnitude of the reference velocity is accelerated to $10$ m/s, the system will switch to a circle trajectory with a constant speed. 

We compare the position error of the lifting-wing quadcopter flying the circle trajectory described above for four conditions: (i) PID velocity controller expressed in \eqref{Eq: vel controller} with feedforward references computed from the differential flatness transform considering aerodynamic forces (PID+DFAF), (ii) PD velocity controller with feedforward references considering aerodynamic forces (PD+DFAF), (iii) PID velocity controller with feedforward references without considering aerodynamic forces (PID+DF), and (iv)  PD velocity controller with feedforward references without considering aerodynamic forces (PD+DF). {We have also tried to remove the angular velocity feedforward. As a result, the aircraft deviated significantly from the trajectory when the speed is increased to about 9m/s. Therefore, this flight data is not included in the comparison.}

Fig. \ref{Fig: Circular Trajectory} shows experimental results for tracking a circular trajectory reference on conditions (ii) and (iv). The velocity has a low-frequency sine wave, caused by the state estimator, because the update frequency of GPS is only 10Hz, which brings a considerable delay to the position estimation. The altitude tracking error is far less than that of the horizontal position, because, on the one hand, the Lidar provides higher measurement accuracy than GPS, and on the other hand, the altitude command is constant. The tracking performance statistics for four conditions are summarized in Fig. \ref{Fig: Compare}. From these statistics, it can be observed that the trajectory tracking performance has been improved significantly when considering aerodynamic forces and the integral can reduce tracking error, but its effect is very limited. The $E_p$ of conditions (i) and (ii) are about half of that of (iii) and (iv). Fig. \ref{Fig: circle_feedforward} shows states of the vehicle agree well with the feedforward commands, demonstrating that the simplified aerodynamic model \eqref{Eq: aero forces} has fairly good fitting accuracy.

\subsection{Lemniscate Trajectory}
The Lemniscate trajectory is expressed as
\begin{equation}
	\mathbf p_{\rm ref} = \mathbf p_{0}+\left[ r(1-\cos(0.5\omega t))\ r\sin(\omega t)\ 0\right]^{\rm T}.
\end{equation}
Here, we set $r=20$ m, $\omega = 0.33$ rad/s. Fig. \ref{Fig: Lemniscate Trajectory} shows experimental results for tracking a Lemniscate of Gerono with an increasing speed from 0 m/s to 8 m/s on conditions (ii) and (iv) with the same aerodynamic coefficients in the circle trajectory. It also can be observed that the trajectory tracking performance has been improved significantly when considering aerodynamic forces, with the $E_p$ reduced by half.

\section{CONCLUSION}  \label{Sec: conclusion}
In this paper, we propose an effective unified control law to track agile trajectories for lifting-wing quadcopters. First, we derive the differential flatness of the lifting-wing quadcopters, which can transform the reference trajectory into the velocity, acceleration, thrust, attitude, and angular velocity commands. Then, the proposed controller combines these feedforward commands and feedback outputs to accurately track agile trajectories. The outdoor experiment results show that the proposed control law can reduce the tracking error by half.
 



\bibliographystyle{IEEEtran}
\bibliography{IEEEabrv,mylib_abrv}

	\end{sloppypar}
\end{document}